\documentclass[10pt,journal,compsoc]{IEEEtran}
\ifCLASSOPTIONcompsoc
  \usepackage[nocompress]{cite}
\else
  \usepackage{cite}
\fi
\ifCLASSINFOpdf
   \usepackage[pdftex]{graphicx}
\else
   \usepackage[dvips]{graphicx}
\fi
\usepackage{amsmath}
\usepackage{array}
\usepackage{multirow}

\usepackage{bm}

\usepackage{changes}

\usepackage{textcomp}

\begin{document}
%
% paper title
% Titles are generally capitalized except for words such as a, an, and, as,
% at, but, by, for, in, nor, of, on, or, the, to and up, which are usually
% not capitalized unless they are the first or last word of the title.
% Linebreaks \\ can be used within to get better formatting as desired.
% Do not put math or special symbols in the title.
\title{Automated Pain Detection from Facial Expressions using FACS: A Review}
%
%
% author names and IEEE memberships
% note positions of commas and nonbreaking spaces ( ~ ) LaTeX will not break
% a structure at a ~ so this keeps an author's name from being broken across
% two lines.
% use \thanks{} to gain access to the first footnote area
% a separate \thanks must be used for each paragraph as LaTeX2e's \thanks
% was not built to handle multiple paragraphs
%
%
%\IEEEcompsocitemizethanks is a special \thanks that produces the bulleted
% lists the Computer Society journals use for "first footnote" author
% affiliations. Use \IEEEcompsocthanksitem which works much like \item
% for each affiliation group. When not in compsoc mode,
% \IEEEcompsocitemizethanks becomes like \thanks and
% \IEEEcompsocthanksitem becomes a line break with idention. This
% facilitates dual compilation, although admittedly the differences in the
% desired content of \author between the different types of papers makes a
% one-size-fits-all approach a daunting prospect. For instance, compsoc
% journal papers have the author affiliations above the "Manuscript
% received ..."  text while in non-compsoc journals this is reversed. Sigh.

\author{Zhanli~Chen,~\IEEEmembership{Student Member,~IEEE,}
        Rashid~Ansari,~\IEEEmembership{Fellow,~IEEE,}
        and~Diana J. Wilkie, ~\IEEEmembership{Fellow,~AAN,} % <-this % stops a space
\IEEEcompsocitemizethanks{
\IEEEcompsocthanksitem Z. Chen and R. Ansari are with the Department
of Electrical and Computer Engineering, University of Illinois at Chicago \protect\\
% note need leading \protect in front of \\ to get a newline within \thanks as
% \\ is fragile and will error, could use \hfil\break instead.
E-mail: zchen35@uic.edu, ransari@uic.edu
\IEEEcompsocthanksitem D. Wilkie is with the Department of Biobehavioral Nursing, University of Florida \protect\\
E-mail : diwilkie@ufl.edu
}% <-this % stops an unwanted space
\thanks{Manuscript received May 28, 2018;}}
\IEEEtitleabstractindextext{%
\begin{abstract}
Facial pain expression is an important modality for assessing pain, especially when the patient's verbal ability to communicate is impaired. The facial muscle-based action units (AUs), which are defined by the Facial Action Coding System (FACS), have been widely studied and are highly reliable as a method for detecting facial expressions (FE) including valid detection of pain. Unfortunately, FACS coding by humans is a very time-consuming task that makes its clinical use prohibitive. Significant progress on automated facial expression recognition (AFER) have led to its numerous successful applications in FACS-based affective computing problems. However, only a handful of studies have been reported on automated pain detection (APD), and its application in clinical settings is still far from a reality. In this paper, we review the progress in research that has contributed to automated pain detection, with focus on 1) the framework-level similarity between spontaneous AFER and APD problems; 2) the evolution of system design including the recent development of deep learning methods; 3) the strategies \added{and considerations} in developing a FACS-based pain detection framework from \added{existing} research; and 4) \added {introduction of the most relevant databases that are available for AFER and APD studies.} We attempt to present key considerations in extending a general AFER framework to an APD framework \added {in clinical settings.} In addition, the performance metrics are also highlighted in evaluating an AFER or an APD system.

%The recent progress on computer vision and machine learning (CVML) techniques have led to numerous successful applications on FACS based affective computing problems. However only a handful research has been reported on automated pain for  Nowadays, pain evaluation still relies on self-report from patients., an AFER system that reliably detecting pain-related AUs would be highly beneficial for efficient and practical pain monitoring. However there is only a few research on automated FACS based systems for pain detection that focus on very limited type of pain so far, the reason is majorly due to the difficulty in data acquisition and variety of pain characteristics. In this paper, we give a review on existing progress on the development about automated pain detection, with focus on 1) techniques employed in state of the art AFER systems, 2) the challenges from establishing pain oriented video dataset under clinical settings and 3) strategies in developing FACS based pain detection framework from individual research. We attempt to present common problems to be considered in developing an automated system for pain detection from facial expressions under clinical settings and In addition, the measure metrics are also highlighted in performance evaluation of the automated system with respect to patients and observers (care givers).

%generic pain detector different type of pain acute pain chronic pain

\end{abstract}

% Note that keywords are not normally used for peerreview papers.
\begin{IEEEkeywords}
FACS, Action Unit, Automated Pain Detection, Facial Action Coding System, Automated Facial Expression Recognition, Deep Learning
\end{IEEEkeywords}}

% make the title area
\maketitle

%\linenumbers

\section{Introduction}
% facial expression is a strong indicator of pain
\IEEEPARstart{P}{ain} assessment is essential to providing proper patient care and \added{assessing} its efficacy under clinical settings. A patients' self-report is commonly used as the 'gold standard' to report pain through manual pain measurement tools, including the verbal numerical scale (VNS) and visual analogue scale (VAS). However, human sensing and judgement of pain is subjective and the scale report may vary significantly among individuals. Behavioral observation of a patient, in particular the use of facial expression, as a key behavioral indicator of pain, has been identified as an important modality for assessing pain \cite{williams2002facial} , especially when the patient's ability to communicate pain is impaired\cite{abu2000challenge}. Patients who are dying, intellectually disabled \cite{mcguire2010chronic}, critically ill and sedated \cite{payen2001assessing}\cite{arif2010facial}, or have dementia \cite{manfredi2003pain}, head and neck cancer, or brain metastasis \cite{herr2006pain}\cite{hadjistavropoulos2007interdisciplinary}
\cite{puntillo2004pain} are particularly vulnerable and in need of technology that could provide reliable and valid alerts about their pain to busy clinicians. The American Society for Pain Management Nursing (ASPMN), in its position statement on pain assessment in the nonverbal patient \cite{herr2006pain}, describes a hierarchy of pain assessment in which the observation of behavior including facial expressions is noted to be a valid approach to pain assessment. McGuire \emph{et al} \cite{mcguire2010chronic} concluded that pain in the intellectual disability population “may be under-recognized and under-treated, especially in those with impaired capacity to communicate about their pain”. A study of patients undergoing procedural pain \cite{puntillo2004pain} showed a strong relationship between procedural pain and behavioral responses and it identified specific procedural pain behaviors that included facial expressions of grimacing, wincing, and shutting of eyes. It was relatively rare for facial expressions to be absent during a painful procedure. Findings by Payen \emph{et al} \cite{payen2001assessing} strongly support the use of facial expression as a pain indicator in critically ill sedated patients. A study of pain assessment \cite{manfredi2003pain} in elderly patients with severe dementia provides evidence that patients with pain show more pronounced behavioral response compared with patients without pain and note that clinician observations of facial expressions are accurate means for assessing the presence of pain in patients unable to communicate verbally because of advanced dementia. Research has shown that facial expressions can provide reliable measures of pain across human lifespan and culture varieties \cite{williams2002facial} and there is also good consistency of facial expressions corresponding to painful stimuli. Assessment of facial expression of pain not only brings added value when verbal report is available but also served as a key behavior indicator for pain in the scenario of non-communicating patients.

% calls for automated pain detection system. Pain related AUs are consistent To what extent can work on facial expression of emotion be extended to pain

Detailed coding of facial activity provides a mechanism for understanding of different parameters of pain that are unavailable in self-report measures \cite{craig1992facial}. The Facial Action Coding System (FACS)\cite{ekman1978manual}\cite{ekman2002facial}\cite{wilkie1995facial}, in which pain behaviors are observed while a patient undergoes a series of structured activities \cite{keefe1982development}, have been found to be useful in assessing pain among seniors with and without cognitive impairments in research. FACS provides a objective description of facial expressions with $30$ action units(AU) based on the unique changes produced by muscular movements. Several studies using FACS conducted by Prkachin \emph{et al} \cite{craig1980ontogenetic}\cite{craig1992facial} identify six action units \added{associated with pain:} including brow lowering (AU4), cheekraising/lid-tightening (AU6/7), nose wrinkle and upper lip raise (AU9 and AU10), and eye closing (AU43).  Prkachin \emph{et al} \cite{prkachin2008structure} developed the Prkachin and Solomon Pain Intensity (PSPI) metric based on this observation, which is a $16$-level scale \added {based on the contribution of individual intensity} of pain related AUs and is defined as:
\begin{equation*}
    Pain=AU4+max(AU6,AU7)+max(AU9,AU10)+AU43
\end{equation*}

The list of pain-related AUs has been further expanded in more extensive research \cite{williams2002facial} to include lip corner puller (AU12), lip stretch (AU20), lips part (AU25), jaw drop (AU26) and mouth strectch (AU27). The collection of $11$ AUs that occur singly or in combination are considered as core Action Units to specify the prototype of pain, both acute and chronic types, in most FACS based pain study. However, full FACS coding procedure is laborious, and it requires $100$ units of time coding for every unit of real-time, which is unlikely to be useful in busy clinical settings \cite{hadjistavropoulos2007interdisciplinary}. The development of an automated system that identifies a configuration of facial actions associated with pain would be a significant innovation for enhanced patient care and clinical practice efficiency.

Over the past two decades, significant progress has been achieved in automated facial expression recognition. Advances in computer vision (CV) techniques has established robust facial image perception and modeling under naturalistic environment, and state-of-the-art machine learning (ML) methods have been adopted in the research on spontaneous facial expression recognition. These developments in Computer Vision and Machine Learning (CVML) have greatly motivated researchers seeking practical solutions to various affect-related problems, where the term "affect" in psychology is associated experience of feeling or emotion. However, only a handful of studies have focused on automated pain detection from facial expressions.  A literature survey conducted from 1982 to June 2014 with keywords "facial action coding system pain" by Rojo \emph{et al} \cite{rojo2015pain} found only 26 relevant references. Prkachian notes that “information technology-based systems are currently relatively primitive, but there is little doubt that their use in assessing pain expression is feasible and their capacities will become more powerful with further research.” \cite{prkachin2009assessing}. However, the challenges impeding the progress on clinical applications of automated pain detection arise largely from the following aspects, 1) considerable difficulty on facial pain data collection from critically ill patient, 2) insufficient well-annotated data of pain in existing datasets, 3) imbalanced dataset with a limited portion of positive (pain) samples, 4) rigid head motions and low AU/pain intensity in most spontaneous facial expression datasets, 5) inadequate methodology that uses CVML models to mimick human observers making pain assessments. As a result, most existing APD research has access to a very limited selection of pain datasets and has mainly focused on acute pain elicited by a single type of stimuli, which is typically in a laboratory environment \cite{hammal2014towards}. On the other hand, human ability to perceive pain in others is significantly developed by the age of five to six and the sensitivity to more subtle facial signs of pain generally increases with age \cite{deyo2004development}, \added{which suggests that an APD system mimicking human decision procedures of accessing pain is within the scope of present CVML techniques}.

Identifying pain from facial expressions can be viewed as an extension to the spontaneous facial expression recognition problem, which is a more general research area for affect-related learning. An APD framework can be built naturally atop an AFER system as they can share most modules of system design. In addition, pain facial action units are also associated with other affects, including disgust, sadness, fear or anger \cite{williams2002facial}.   In this survey, we present a systematic review of APD development, with focus on how it can benefit from the advances in AFER research as well as from the progress in deep learning technology. In particular, we refer AFER/APD methods using only classic CVML techniques, that are not deep learning related, as conventional approaches. The paper is orgnized as follows: Section 2 summarizes important facial expression datasets for AFER and APD. Section 3 reviews an AFER framework in terms of block customization, which serves as the basis for describing APD research.  Section 4 presents a systematic overview of the APD development that builds on the available body of AFER research. Section 5 discusses the evolution of AFER and APD with progress on deep learning. Finally, Section 6 summarizes the state-of-the-art work and concludes with discussion on future work of APD in clinical settings.

\section{Publicly Accessible Facial Expression Datasets}
Availability of representative datasets is the fundamental basis for training and testing an AFER system. The performance evaluation of and comparison among different AFER approaches would not be fair or meaningful without context of the datasets used in the evaluation. Before proceeding to an overview of specific AFER and APD approaches, we will present an introduction to selected publicly available facial expression datasets in this section. The datasets are grouped depending on the context of the problem and present them in a chronological order of availability. These datasets were all used in the studies to be covered in this paper.
\subsection{Classic Facial Expression Datasets for Conventional AFER}

\paragraph*{\textbf{MMI (2005)}} The MMI dataset \cite{pantic2005web} contains $740$ static images and $780$ video sequences captured in frontal and profile views from 19 subjects of students and research staff. Two thirds of image samples and frames of $169$ video sequences have been annotated by two certified FACS coders for $30$ AUs. The temporal phases of an AU event (onset, apex, offset) are also coded for the $169$ sequences. The MMI is a dataset of posed (not spontaneous) facial expressions under controlled settings.

\paragraph*{\textbf{RU-FACS (2005)}} The RU-FACS (or \textbf{M3}) dataset \cite{bartlett2006automatic} captures spontaneous facial expressions from video-recorded interviews of $100$ young adults of varying ethnicity. Each video segment lasts about $2$ mins on average, where the head pose is close to frontal view with small to a moderate out-of-plane rotation. 33 subjects have been coded by two certified FACS coder for $19$ AUs.

\paragraph*{\textbf{CK+ (2010)}} The extended Cohn-Kanade dataset \cite{lucey2010extended} was constructed upon the Cohn and Kanade\textquotesingle s DFAT-504 dataset [\textbf{CK (2000)}] \cite{kanade2000comprehensive} contains $593$ videos captured under controlled conditions from $123$ subjects with various ethnic backgrounds. Each brief video segment (approximately $20$ frames on average) begins with a neutral expression and finishes at the apex phase. The entire dataset is annotated for $30$ AUs at frame-level. The CK+ dataset has been made available since $2010$ and is widely used for prototype and benchmark AFER systems.

\paragraph*{\textbf{GEMEP-FERA (2011)}} The GEMEP-FERA dataset \cite{valstar2012meta} is a fraction of the geneva multimodal emotion portrayal (GEMEP) corpus \cite{banziger2010introducing} employed as a benchmark dataset for the first facial expression recognition challenge (FERA2011). The original GEMEP corpus consists of over 7000 portrayals with $18$ posed facial expressions portrayed by 10 actors. The GEMEP-FERA dataset employs $158$ portrayals for AU detection sub-challenge and $289$ portrayals for emotion sub-challenge. Ground truth is provided as frame-by-frame AU coding for $12$ AUs and event coding of $5$ discrete emotions.

\paragraph*{\textbf{GFT (2012)}} The Sayette Group Formation Task (GFT) dataset \cite{sayette2012alcohol} captures spontaneous facial expressions from multiple interacting participants in a psychological study to evaluate socioemotional effects of alcohol. The dataset include $172,800$ video frames from $96$ participants in $32$ three-person groups \cite{girard2017sayette}. Multiple FACS coders are recruited for binary occurrence coding for $19$ AUs on frame-level, and $5$ AUs are selected for intensity coding (level: A-E). The baseline results are presented in a recent publication \cite{girard2017sayette} for $10$ AUs using maximal-margin framework (linear SVM) and deep learning framework (AlexNet).

\subsection{Facial Expression Dataset for Automated Pain Detection}
Establishing a pain-oriented facial expression dataset is much more challenging than establishing a general AFER dataset.
\paragraph*{\textbf{UNBC-McMaster (2011)}} The UNBC-McMaster Shoulder Pain Expression Archive Dataset \cite{lucey2011painful} is the only publicly available spontaneous facial expression dataset targeting pain. It contains $48,398$ FACS coded frames in $200$ video sequences. The video segments are captured from patients suffering from shoulder pain and spontaneous facial expressions are triggered by moving their affected and unaffected limbs. All frames are coded by certified FACS coders for $10$ single pain-related AUs and the frame-level pain score is rated by the Prkachin and Solomon Pain Intensity (PSPI). A sequence-level pain label is assigned by self-reported VAS and Observer-rated Pain Intensity (OPI). In addition, $66$-point facial landmarks from Active Appearance Model (AAM) are also provided for each frame to facilitate the development of a user-customized AFER system. It is the only publicly available pain-oriented facial expression video dataset in which the spontaneous facial expressions are evoked solely by acute pain.

%Video Frames/ Sequnce
%AU classification
%Pain Detection: UNBC-McMaster EMOPAIN
%Deep Learning Benchmarking Dataset
%Psychology environment
%
%One milestone on the APD research is the establishment of the UNBC-McMaster Shoulder Pain Archive
%
%pain label VAS OPI PSPI binary
%
%Population Pain Type

\subsection{Facial Expression Datasets for Deep Learning Based AFER}
\paragraph*{\textbf{SEMAINE (2012)}} As part of a project: Sustained Emotionally colored Machine-human Interaction using Nonverbal Expression, the SEMAINE dataset \cite{mckeown2012semaine} provides high-quality, multimodal recording to study social signals occurring during interaction between human beings and virtual human avatars. The video frames were recorded from $150$ participants of varying age and partitioned into subsets for training ($48,000$), development ($45,000$), and testing ($37,695$). Spontaneous facial expressions were annotated by three certified FACS coder for $6$ AUs at frame-level. SEMAINE was selected as a benchmark dataset in FERA 2015 challenge \cite{valstar2015fera}.

\paragraph*{\textbf{DISFA (2013)}} The Denver Intensity of Spontaneous Facial Action (DISFA) Database  \cite{mavadati2013disfa} captures spontaneous facial expressions elicited by viewing a 4-mins video clip from $27$ young adults via stereo video recording under uniform illumination condition. $4845$ frames were recorded for each participant and coded by a single FACS coder for 12 AUs (5 upper face AUs and 7 lower face AUs) with $6$ (0-5) level of intensities. The number of events (from onset to offset) and the number of frames for each intensity level were also recorded. A second FACS coder was recruited for annotating video from $10$ randomly selected participants and resulted in a high interobserver reliability (ICC) ranging from $0.80$ to $0.94$.

\paragraph*{\textbf{BP4D (2014)}} The Binghamton-Pittsburgh 4D Spontaneous Expression Database (BP4D) \cite{zhang2014bp4d} contains spontaneous facial expression elicited by well-validated emotion instructions associated with eight tasks, which are captured from a diverse group of young adults. A total of $639,224$ frames are produced from forty-one participants and annotated for $27$ AUs by FACS certified coders in a 20-second segment interval. The video is available in both 3D and 2D, where geometric \added{features are represented} by a 83-point 3D temporal deformable shape model (3D-TDSM) and a 49-point 2D shape model tracked by a constrained local model (CLM). BP4D was also selected as a benchmark dataset in FERA 2015 \cite{valstar2015fera}.

\section{FACS Based AU Detection}
\subsection{FACS Coding}
The message-based and sign-based judgements are two major approaches in studying facial behavior \cite{ekman1969repertoire}. While message-based judgements make inferences about emotions that underlying facial expressions, e.g. basic emotions or pain, sign-based judgments focus on facial expressions itself by describing the surface of behavior created by muscular movement. FACS provides a standard technique for measuring facial movement by specifying $12$ action units in the upper face and $18$ in the lower face. The action units are defined on the anatomical basis of muscular movement that are common to all people. Depending on the research question, observers commonly use two schemes to score AUs. The comprehensive scheme codes all AUs in a chosen video segment, and the selective scheme only codes predetermined AUs. The reliability of AU scoring, also referred to as inter-observer agreement, is assessed by considering four criteria:  the occurrence of an AU, the intensity of an AU, the AU temporal precision defined as onset, apex, offset and neutral stage, and the aggregates that code certain AUs in combination. While the occurrence of AU is most widely used criterion to evaluate an AFER system, the latter three are evaluated depending on specific research problems. For example, an event can be decomposed into a set of action units that overlap in time, thus a link can be created between coding of pain via message-based mechanism and FACS-based system via AU aggregates coding.
%The coefficient kappa is used to access such reliability to control for agreement due to chance.
% pain can be evaluated either in direct way or indirect way through sign based judgement

\subsection{General AFER Framework}
Over the past two decades, AFER research has generally adopted a framework that is comprised of four core modules: face detection, face alignment, feature extraction and facial expression classification.
\subsubsection{Face Detection}
The face detection module locates face regions with a rectangular bounding box in a video frame so that only localized patches are adopted for further processing. One example of a state-of-the-art face detector is real-time Viola-Jones detection framework \cite{viola2001rapid} that employs the Adaboost learning algorithm in a cascade structure for classification. The ready-to-use Viola-Jones detector is capable of robustly detecting multiple faces over a wide range of poses and illumination conditions, as a result of which it is widely employed in most AFER research in practice. An extensive review for available face detection approaches is presented in \cite{zhang2010survey}.

\subsubsection{Face Alignment in 2D}
The face alignment module tracks a set of fiducial points on the detected face area in every video frame. The fiducial points are typically defined along the cues that best describe features of a face, including jawline, brows, eyes, nose, and lips. The set of fiducial points is also referred to as the shape model when a parametric model is applied to human face. While facial expressions produce the non-rigid facial surface change, head pose rotations account for rigid motions. Although rigid motions are negligible in controlled experimental settings for posed facial expression dataset, occurrence of spontaneous facial expressions in natural environment is frequently accompanied by obvious rigid motions, where challenges arise for accurate feature extraction in the next step. Rigid motion normalization can be embedded within an alignment framework via similarity or affine transform. Alternatively, an aligned face model can be rotated and frontalized with the estimated head pose parameters and registered to a front view mean face template with neutral facial expression.

Faces can be modelled as deformable objects which vary in terms of shape and appearance. The active appearance model (AAM) \cite{cootes2001active}\cite{baker2004lucas} is one of the most commonly used face alignment method, where the optimization process relies on jointly fitting a shape model and an appearance model via the steepest decent algorithm over the holistic face texture. The original AAM is subject-dependent where the model is tuned specifically to the environmental conditions of the training dataset. This, however, may cause performance degradation on face images of an 'unseen' subject. This problem can be addressed by using the constrained local model (CLM) \cite{cristinacce2006feature}\cite{cristinacce2008automatic}, which is comprised of a shape model and an appearance model similar to AAM. It only utilizes rectangular local texture patches around the shape feature points to optimize the shape model through a non-linear constrained local search. The appearance templates in CLM are represented by a small set of parameters that can be effectively used to fit unseen face images. Therefore a CLM framework can be conveniently modified as a subject-independent model \cite{asthana2013robust} for generic face alignment applications. The result in \cite{chew2012pursuit} indicates that CLM is generally more \added{computationally} efficient than AAM but it is slightly outperformed by AAM when rigid motions are presented.

Follow-up research led to continuous improvement to the family of AAM and CLM methods in terms of fitting accuracy and computational efficiency, which involvs adaptation to the fitting methodology \cite{tzimiropoulos2017fast}, feeding rich annotated training data in unconstrained conditions \cite{zhu2012face}, combination with new appearance descriptors \cite{xiong2013supervised}, and fine tuning with cascaded regression techniques \cite{asthana2014incremental}. Such evolutions have boosted the performance of face alignment to handle illumination variations, out-of-plane rigid motions, as well as partial occlusions. It is worth to noting that typical AU coding scenario is at close-to-front ($\pm30^{\circ}$) view, which can be adequately handled by recent 2D alignment methods. Although 3D alignments provide much denser geometric modeling as well as depth information, computational inefficiency and difficulty in data acquisition makes it currently infeasible in many practical AFER and APD applications. Therefore we shall focus on automated pain detection from 2D facial expression recognition in this paper. For an adequate survey on facial expression modeling in 3D, readers are referred to the survey \cite{sandbach2012static}.

 %rigid motion non rigid motion
 %AAM person specific CLM generic
 %
 %shape geometry appearance texture info

\subsubsection{Feature extraction}
Facial expressions are revealed by non-rigid facial muscular movement. A feature extraction module is added between raw input pixels and classifiers and serves as an important interpreter to extract relevant non-rigid information for manifesting Action Units from computer vision perceptions, which is highly customizable in the framework structure with a rich option of feature descriptors. Features can be categorized into geometric-based and appearance-based. Geometric features extract geometric measurements related to coordinates of fiducial points, while appearance features are derived from pixel intensities. Predefined distance, curvatures and angles from displacement of the fiducial points were an intuitive way to extract geometric features in early AFER research \cite{tian2001recognizing}\cite{pantic2006dynamics}. Static geometric measurements can be used as midlevel representations to extract dynamic features by encoding with temporal information for a more detailed description of neuromuscular facial activities \cite{valstar2012fully}.  Geometric features are typically low in dimension and simple to compute and focus on describing the deformation of the conspicuous feature cues. However, geometric features are insufficient to model muscular actions off the feature cues and they are sensitive to registration errors from shape alignment. Therefore geometric features are typically applied in conjunction with other feature descriptors in the spontaneous AFER scenario.

On the other hand, appearance features could be extracted directly from the alignment modality \cite{ashraf2009painful}, either by similarity normalized appearance (S-APP) that normalizes the rigid motion or by canonical appearance (C-APP) that warps the texture to the mean shape template. The resulting appearance descriptor is then concatenated in a high-dimensional feature vector based on raw pixel intensity for classification. In this case, facial-expression-related muscular activities are encoded together with a large portion of person specific features without discrimination, which potentially attenuates the discriminating power of the feature vector towards target AUs. Subtle texture changes like deepening the nasolabial furrow, raising the nostril wing, cheek rising, pushing the infraorbital triangle up, provide important reference for coding of certain AUs \cite{ekman1978manual}. These non-rigid motions are nonstructural, which are infeasible to be effectively described by raw geometric or pixel-based features. Descriptors developed for object detection are introduced in AFER research in order to exploit the features more in depth. These descriptors detect local intensity variations as feature points, edges, phase or magnitudes and assemble the low level appearance features into high-level feature representations, from which a meaningful classification can be made. A series of studies \cite{bartlett2006automatic}\cite{wu2010facial}\cite{littlewort2011computer}\cite{wu2012multilayer} have applied a Gabor filter bank to extract features in 8 directions and 9 spatial frequencies and the output magnitudes are concatenated into a single feature vector for AU classification. The local binary pattern (LBP) encodes local intensity variation in the neighbourhood and as an extension to LBP, the local phase quantization (LPQ) captures local phase information from Fourier coefficients, and a histogram is assembled from the local descriptors as a high-level feature representation. If the facial video volume is sliced by three orthogonal planes (TOP), both LBP and LPQ static feature descriptors are then expanded to encode temporal information from $x-t$ and $y-t$ planes, which are known as LBP-TOP and LPQ-TOP \cite{jiang2011action}.   A detailed survey of the application of the family of LBP descritpors to facial image analysis can be found in \cite{huang2011local}. More geometric-invariant feature descriptors have been successfully applied to AFER including the histogram of oriented gradient (HOG) \cite{chew2012pursuit} and scale-invariant feature transform (SIFT) \cite{girard2015spontaneous}. The appearance descriptors can also be associated with an image pyramid representation to extract feature with multiple scales and resolutions \cite{dhall2011emotion}\cite{sun2014combining}.

In addition to extracting static features from a single frame, it is also beneficial to extract dynamic features comprised of spatially repetive, time-varying patterns \cite{chetverikov2005brief} to model motion-based information from an image sequence.  Koelstra \emph{et al } \cite{koelstra2010dynamic} employ two representation methods, motion history image (MHI) and free-form deformation (FFD), to derive non-rigid motions from consecutive frames. A MHI  descritpor computes a motion vector field from a set of weighted binary difference images within a predefined time window, while a FFD descriptor outlines non-rigid motions via a B-spline interpolation that computes local variations of a control point lattice. A quad-tree decomposition method is then applied to decompose the face region in a non-uniform grid representation, where most grids are placed on areas with high motion response. The quad-tree decomposition is performed in three orthogonal planes to extract features in terms of magnitude, horizontal motion, and vertical motions. This work \cite{koelstra2010dynamic} provides an example on feature customization in terms of geometric and appearance feature combination, dynamic feature extraction by utilizing temporal information, as well as a balance between global and local feature extraction. Each AU is activated by a combination of facial muscles, which can be geometrically located in a specific region on the face. Appearance features extracted from corresponding region of interests (ROI) bounded by geometric features potentially help to reduce the feature dimensionality and boost computation efficiency. Hamm \emph{et al} \cite{hamm2011automated} divided a face image into $14$ rectangular regions based on geometric landmarks’ layout, from which a $72$-dimensional Gabor response is pooled to generate a histogram based feature representation. If multiple feature descriptors are employed in an AFER framework, a feature fusion problem should also be considered. Wu \emph{et al } \cite{wu2012multilayer} evaluate the performance multi-layer architectures for feature extraction involving a combination of two texture descriptors, the Gabor Energy Filter (GEF) and LBP. However, a more common way for feature fusion is to perform it in the classification stage, which will be addressed in the next section.

%Dynamic feature Temporal Local patch(gemometric bounded region segmentation Feature fusing
%
%49 points shape mode without jawline or 66 points shape model include jawline
%pyramid cascade
%coarse to fine search
%global local dynamic
%grid
%DAISY
%feature detector
%
%Appearance Gabor HOG LBP LPQ LPQ-TOP SIFT DAISY high dimensional, hard to find an efficient mapping between AU high level transition and texture change
%Combination of shape and appearance
%fusing of multiple feature descriptor
%multi level on top of
%In the language of neural networks, current expression recognition architectures are predominantly “shallow” or have a “single hidden layer”: The input layer is the set of raw pixel intensities, the hidden layer is a bank of non-linear texture filters, and the output layer is a single layer classifier. Shallow architectures are currently
%
%local patch
%hand crafted

\subsubsection{AU classification and regression}
FACS-based AFER studies typically focus on two core problems in Action Unit recognition: 1) which AUs are displayed in the input images or videos, and 2) how to measure the intensity of the observed AUs. Conventional AFER research frequently treats the detection of AU occurrence as a binary classification problem, where a binary classifier is trained for each of the target AU independently, and a separate set of classifiers or regressors are trained to discriminate AU intensity of different levels. Static classification approaches serve as the basis for AU recognition, which rely on features extracted from one static image frame. However, psychological studies \cite{bassili1979emotion} suggest that facial behavior can be more reliably recognized from an image sequence than from a still image. Hence dynamic approaches include additional analysis steps on static predictions by utilizing temporal information from adjacent frames to improve reliability of frame-level AU estimations or perform high-level AU event coding across the video sequences.

\paragraph*{\textbf{Static Approach}}
Support vector machine (SVM) and boosting-based classifiers are commonly employed for the AU classification task. A binary support vector machine seeks a subset from training data known as support vectors, which defines a hyperplane that maximizes the margin between the hyperplane and closest points of both classes. In practice, the feature spaces can be lifted to a higher-dimensional feature space using \emph{kernels} \cite{szeliski2010computer}, where linear and radial basis function (RBF) kernels are frequently used in SVM settings for AU classification. In fact, SVM was used as part of the baseline system in most facial expression datasets \cite{bartlett2006automatic}\cite{lucey2010extended}\cite{lucey2011painful}\cite{mavadati2013disfa}\cite{girard2017sayette}. A boosting-based approach learns a set of weak binary classifiers in a sequential order, where mis-classified samples are assigned higher weights and are considered as 'hard' problems to be handled with the next weak classifier. The final decision is made by a strong classifier that is generated from a linear combination of the set of weak classifiers.

Bartlett \emph{et al} \cite{bartlett2006automatic} assessed both SVM (linear and RBF kernels) and Adaboost (up to 200 features selected per AU) classifiers for facial action classification on $20$ AUs, where Gabor wavelet features were extracted from the RU-FACS (spontaneous) and CK (posed) datasets. Chew \emph{et al} \cite{chew2012pursuit} employ a linear SVM to test pixel-based features and more complex appearance features (HOG, Gabor Magnitudes) extracted from multiple datasets (CK+, M3, UNBC-McMaster and GEMEP-FERA) under different face alignment accuracy. The experiment suggests that the more complex appearance descriptors are robust to alignment errors on AU detection, but their advantages are limited under close-to-perfect-alignment. Jiang \emph{et al} \cite{jiang2011action} performed recognition of $23$ upper and lower face AUs on MMI dataset with four different SVM kernel settings, including Linear, Polynomial, RBF and Histogram intersection. The SVM classifiers were trained with a set of features from LBP family with optional \emph{Block} and \emph{Pyramid} extensions, including LBP, LPQ, B-PLBP and B-PLPQ. Experimental results indicated that a SVM with histogram intersection provided slightly better performance than other kernel settings, and the block-pyramid based B-PLPQ feature outperformed other features but at the cost of increased computational complexity. Jeni \emph{et al } \cite{jeni2013continuous} proposed a continuous AU intensity  estimation framework using support vector regression (SVR). Sparse features were learned from local patches via personal mean texture normalization followed by non-negative maxtrix factorization. A $L_2$-loss regularized least-squares SVM (LS-SVM) model was trained for AU intensity regression in $6$ ordinal levels (0-5). The proposed AU intensity estimation system was tested on $14$ AUs on posed data from CK+ as well as on spontaneous data from BP4D for AU12 and AU14.  Chu \emph{et al} \cite{chu2017selective} investigated the problem of personalizing a classifier trained on a generic facial expression dataset with respect to unlabeled person-specific data through a transductive learning method, which is referred to as Selective Transfer Machine (STM). The idea behind STM is to re-weight the samples in the generic training set by minimizing the training and person-specific test distribution mismatch, such that a person-specific SVM classifier can be generated from the reweighted training data. The proposed system was evaluated for cross-subject AU detection on CK+ and RU-FACS, and cross-dataset AU detetion in terms of RU-FACS$\rightarrow$GEMEP-FERA and GFT$\rightarrow$RU-FACS. The experimental results demonstrated that STM consistently outperformed generic SVM on both tasks.

\paragraph*{\textbf{Dynamic Approach}}
A dynamic approach for AU classification takes input from static AU predictions of multiple consecutive frames to perform temporal analysis that targets three problems in general: 1) improving prediction reliability on the current frame using information from past frames, 2) generate segment-level AU detection from frame-level prediction, and 3) locate AU events in a video sequence by modeling temporal phase transition (i.e. onset, apex, offset, and neutral) of an Action Unit.

Koelstra \emph{et al} \cite{koelstra2010dynamic} trained a one-versus-all GentleBoost classifier for each AU in onset or offset phase indepdently with spatiotemporal features derived from Free-Form Deformations (FFD) or Motion History Image (MHI). The outputs of each pair of onset and offset GentleBoost classifier were combined into one AU recognizer through a continuous Hidden Markov Model (HMM). The four-state HMM performed as a temporal filter as well as AU event encoder by modeling temporal phase transition for each AU, where the prior and transition matrices were estimated from training set and the emission matrix was estimated from the outputs of the GentleBoost classifiers. The proposed system was tested for $27$ AUs in MMI and $18$ AUs in CK+. A follow-up study in \cite{valstar2012fully} proposed a hybrid SVM-HMM framework to estimate the temporal phases of an AU event. A one-versus-one multi-class SVM was trained for each of the four temporal phases using  $2520$-dimensional geometric features. The outputs of SVMs were converted to probability measures via a sigmoid function, which were used to estimate the emission matrix of a HMM. Rudovic \emph{et al} noticed the temporal phase of AUs are correlated with their intensities and proposed an approach based on Conditional Ordinal Random Field (CORF) to utilize the ordinal relations embedded in intensity levels. The proposed LAP-KCORF framework \cite{rudovic2012kernel} extends the CORF model by introducing a kernel-induced Laplacian regularizer to the optimization process. The Composite Histogram Intersection (CHI) kernel was selected in the framework settings, which takes the input of LBP features from aligned training images. The LAP-KCORF model was trained for each  AU separately. While neutral and apex phases can be discriminated solely from the ordinal score, the onset and offset phases have to be discriminated from the dynamic features. Experimental results on $9$ upper face AUs drawn from MMI datasets demonstrated the advantages of the ordinal model over the SVM-HMM model on both AU detection and temporal phase estimation.  Hamm \emph{et al} \cite{hamm2011automated} proposed an AFER system to perform dynamic analysis of facial expressions on a private video dataset featuring patients with neuropsychiatric disorders and healthy controls. A one-versus-all GentleBoost classifier was trained for each of the $15$ selected AUs using Histogram of Gabor Features extracted from multiple pre-defined local regions. The dynamic analysis was measured by single and combined AU frequencies and affective flatness and inappropriateness, which was derived from the temporal profiles of AU predictions on frame level.

A complete AU event refers to activation of an AU in terms of its duration and start/stop boundaries. Realizing the event-based detection is more desirable than frame-based detection in many application scenarios in practice, Ding \emph{et al } \cite{ding2016cascade} carried out AU detection on frame-level, segment-level and transition encoding in a sequential order, and integrated the results from three detectors to Cascade of Tasks (COT) architecture to perform AU event coding. The frame-level detector was trained with SVM on SIFT features, where the frame-level detection were used to augment segment-level training data in both feature representation and sample weight. The segment-level data was represented by a Bag of Words (BOW) structure with geometric features. A weighted-margin SVM was employed for segemnt-level detection, where samples (segments) containing many positive frames (scored by frame-level detector) were associated with higher weights. Two transition detectors was then trained to refine the onset and offset boundaries from the detected AU segments using the segment-level features. The scores for segment-level prediction and transition predictions were linearly combined to produce event prediction of AU. The start and end frames of an AU event were determined based on the highest event score. Multiple events could be scored in a given video sequence using dynamic programming (DP) via a branching-and-bounding strategy. The proposed system was trained and tested for different set of AUs on CK+, RU-FACS, FERA and GFT datasets respectively. Sikka \emph{et al} modeled an affective event as a sequence of discriminative sub-events, and solved the event detection as a weakly supervised problem with ordinal constraints by the proposed Latent Ordinal Model (LOMo) \cite{sikka2016lomo}. A LOMo model seeks a prototypical appearance template to represent each sub-event, and the order of all sub-event permutations is associated with a cost value, where permutations far from the ground truth order was penalized by higher cost. A score function originating from a linear SVM was defined to measure the correlations between a weakly labeled frame (with only sequence-level label) and the templates with consideration of the ordinal cost. The entire model was learned using an algorithm based on stochastic gradient descent (SGD) with respect to a margin hinge loss minimization. The proposed method selected multiple frame-level feature descriptors (SIFT, LBP, geometric and deep-learned features) and was trained and evaluated on four facial expression datasets including CK+ and UNBC-McMaster.

\paragraph*{\textbf{Discover AU Relations}}
The aforementioned methods consider single AU detection as one-versus-all classification problem. There are also efforts that attempt to make better use of FACS by exploiting dependencies and relations among AUs, which are potentially helpful in improving detection on highly skewed AUs in spontaneous facial video datasets. Tong \emph{et al} \cite{tong2007facial} proposed a system based on Dynamic Bayesian Network (DBN) to model relations among different AUs in a probabilistic manner taking into consideration of temporal changes in an AU event. An Adaboost classifier trained on Gabor features was employed as a baseline static AU detector. A Bayesian Network (BN) was used to model the relations of AUs in terms of their co-occurrence and \added {mutually exclusivity}. A DBN network is comprised of interconnected time slices of static BNs, where the state transitions between the consecutive time slices is modeled by a HMM. The structure and parameters of the DBN were obtained by training it offline and AU recognition under temporal evolution was performed online with the trained DBN network. The DBN-based system was learned for $14$ AUs from the CK+ dataset and the validation was further generalized using selected samples from the MMI dataset.  Zhao \emph{et al} \cite{zhao2015joint} formulated patch learning and multi-label learning problems in AU detection as a joint learning framework known as Joint Patch and Multi-label Learning (JPML). Patch learning sought local feature dependencies for each AU, where adaptive patch descriptors were placed around $49$ facial fiducial points to extract $128D$ SIFT features from each patch. Multi-learning exploited strong correlations among AUs from a AU relation matrix that was learned on over $350,000$ FACS annotated frames from multiple datasets. \emph{Positive correlation} and \emph{negative competeition} were defined for relations of likely and rarely co-occurring AU pairs based on the correlation coefficients in the AU relation matrix. The JPML framework integrated the patch learning and multi-label learning into a logistic loss in the form of a patch regularizer and a relational regularizer, which were then jointly solved as an unconstrained problem. The JPML framework was trained and tested on CK+, GPT, BP4D datasets, from which specific active patch regions were identified for $11$ AUs. $8$ positive correlation and $14$ negative competition pairs were discovered among $11$ AUs.

% Binary Action Unit classification, Action Unit intensity prediction, Action Unit combination learning
% rule based when label is not available
% statistical descriptor
%
%svm
%boosting
%intensity model
%dbn
%multi label
%AU intensity
%AU prior multi label
%binary classifier
%
%domain tranfer
%multi model feature classification
%multi level classification
%CERT
%
%
%AU phase modeling
%Event modeling
%late fusion
%Binary Classification
\subsubsection{Performance Metrics}
After training a proposed AFER system on selected datasets, the next step is to evaluate the system performance with proper metrics. The criteria to select measure metrics depends on the type of classification problem, e.g. binary AU classification or multi-level AU intensity estimation.

\emph{Accuracy}, \emph{F1} score and Area Under the receiver operating Characteristic (\emph{AUC}) curve are three most frequently used metrics to evaluate a binary classifier. \emph{Accuracy} is the percentage of correctly classified positive and negative samples. \emph{F1} score is the harmonic mean of between precision (\emph{PR}) and recall (\emph{RC}), so that the two measures are reflected through one score. The metric \emph{AUC} is equal to the probability that a classifier will rank a randomly chosen positive instance higher than a randomly chosen negative one \cite{fawcett2006introduction}.

Evaluation of multi-level intensity classification and regression depends on two types of performance metrics in general. The first type includes the root mean square error (\emph{RMSE}) and mean absolute error (\emph{MAE}) metrics that effectively capture the difference between predicted value and the ground truth. While \emph{MAE} is more intuitive for interpretation, \emph{RMSE} possesses the benefit of penalizing large errors more, and is easier to be manipulated in many mathematical calculations. The second type of metric is correlation based including Pearson's correlation coefficients (\emph{PCC}) and Intraclass correlation (\emph{ICC}), which measure the trend of prediction following the ground truth without considering a potential absolute bias \cite{egede2017fusing}. The \emph{ICC} is similar to \emph{PCC} except the data are centered and scaled using a pooled mean and standard deviation \cite{girard2015spontaneous}, which require raters (e.g. a human coder and an AFER) to provide the same rating without multiplicative difference to achieve agreement on \emph{ICC}. \emph{RMSE/MAE} and \emph{ICC } are often used simultaneously as complimentary metrics in AFER studies, as a low \emph{ICC} is possible with deceptively low  \emph{RMSE/MAE} scores and vice-versa \cite{egede2017fusing}.

Most spontaneous facial expression datasets contain a significantly larger fraction of negative samples than the fraction of positive samples in practice, where the data imbalance can be defined by the skew ratio as $Skew= \frac{negative\,examples} {positive\,examples}$. Jeni \emph{et al} \cite{jeni2013facing} studied the influence of skew on multiple metrics and reported \emph{Accuracy}, \emph{F1}, \emph{Cohen's $\kappa$} and \emph{Krippendorf's $\alpha$} were attenuated by skewed distribution. Although \emph{AUC} was unaffected by skew ratio, it may mask poor performance due to data imbalance. The author suggested to report skew-normalized scores along with original ones to minimize skew-biased performance evaluation. This finding was also taken into account by some studies \cite{tHoser2016deep}\cite{zhao2015joint}\cite{rodriguez2017deep} presented in this paper. It is worth mentioning that all metrics discussed above are also applicable to pain detection and deep learning based AFER approaches. We shall use the same abbreviations (as highlighted) when referring to these metrics in the following sections.

%when observer is involved
%
%Classification Accuracy
%Intensity Measurement
%Correlation between human observer and machine learning
%Metric Skew
%%comercialized productions
%%AU event develop over time

\section{Automated Pain Detection from Facial Expressions}
%relation to AFER system
Present APD studies generally follow two modalities to detect pain from facial expressions. In the indirect way video frames are first encoded with FACS based Action Units and pain is then identified from pain-related AU predictions. Alternatively, a mapping can be learned directly from high-dimensional features to assign pain labels without going through the AU coding procedure. The progress in APD has benefited vastly from advances in AFER techniques which applicable in both modalities, as AFER is the core module in the indirect method and most functional blocks in an AFER framework can be applied within a direct pain detection architecture with minor modifications. In fact, with few exceptions, all APD studies reviewed in this section are linked to some preliminary AFER research. The correlation between the frameworks used in addressing the AFER and APD problems is illustrated in figure 1.
\begin{figure*}
  \centering
  % Requires \usepackage{graphicx}
  \includegraphics[width=7in]{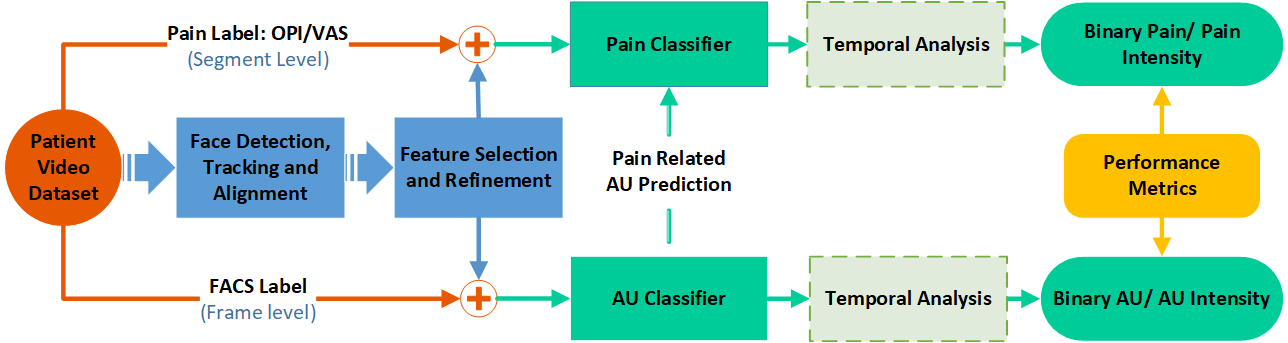}\\
  \caption{Correlation between the frameworks used in addressing the AFER and APD problems}\label{1}
\end{figure*}

Aside from the similarity to the AFER problem, there are additional considerations that should be factored in the design of an APD system. These stem from the nature of the pain data, methods used in labeling pain data, the system performance metrics used, and availability of supplemental pain data from non-facial sensing. For example, pain data may be collected under controlled, spontaneous, or clinical conditions; the ground truth may be obtained from patients' self-report, or certified FACS coder; a pain event may be scored per frame or per segment for the binary pain decision or pain intensity estimation; distinction between acute and chronic pain from onset phase duration \cite{hammal2014towards};  suitable metrics could be used for performance evaluation and for comparison with human coders' decision. Last but not the least, signals from channels other than facial expressions may also help in pain detection. In this section, we will review existing APD research in a case study manner by addressing these considerations.

% NOtES: Previous efforts in a laboratory environment and focus on acute pain, a single modality (e.g., face), and a single type of pain elicitor (e.g., orthopedic [7, 10, 11, 14, 16, 17] or temperature probe [12, 13, 18]). However, pain can have many possible causes (e.g., spinal stenosis, rotator cuff tear, cardiac conditions, or chronic illnesses such as sickle-cell disease) that may produce multiple signals both within and across modalities (e.g. face, body, and voice). Pain can be acute, as a result of injury (e.g., touching a burning ember), or chronic (e.g., (pain thatpersists after an injury heals). Behavioral pain indicators may well differ between acute and chronic pain. Acute pain is likely associated with faster onset. Presence of other people and context further influence the experience and expression of pain. Pain is thus a complex subjective experience that may have varied indicators, which influence its communication to others.

%The systematic overview of automated pain detection mainly concerns five aspects, a)context of problem, b)how facial expression is recognized, c)how pain is identified including temproal dynamic modeling d)dataset  and e) measure metrics and performance evaluation.
%Two task Binary Pain Pain Intensity
%Temproal resolution: frame based event based
%Pain type acute pain chronic pain
%Method: direct indirect

\subsection{Genuine Pain Vs. Posed Pain}
A pioneering APD study examined the problem of distinguishing posed pain from genuine pain through AU coding \cite{littlewort2007faces}\cite{littlewort2009automatic}. Facial expressions induced by posed pain and genuine pain are mediated by two neural pathway and thus have subtle differences in terms of muscular movements and dynamics. The system implementation was comprised of a two-level machine learning module. The first stage solved a general AFER problem by training independant linear SVM binary classifiers for a list of $18$ single AUs and $2$ AU combinations about prototypical expression of fear and distress. Features were extracted by Gabor filters and selected by Adaboost. Differences of Z-scores  were computed for three pain conditions consisting of real pain, fake pain, and absence of pain (baseline). AU 1, 4, and the distress brow combination were considered statistically significant for real vs. fake pain discrimination. Window-based statistics and event-based statistics were obtained from the first stage of AU prediction, \added {and a second layer non-linear SVM was trained to discriminate genuine pain from faked pain.} The AFER system was trained on 3 datasets (2 posed and 1 spontaneous), while the pain data were collected from $26$ subjects subjected to cold pressure pain. The system was evaluated by \emph{AUC} and \emph{2FAC}, and the reported performance found to be superior to naive human judges. Although accuracy of individual AU classifier is still below that of human coders, this is among the first efforts to extend AFER to APD study.

\subsection{Binary Pain Classification}
Following the first study based on UNBC-McMaster dataset, \cite{ashraf2009painful} proposed a framework to detect pain at both frame and sequence level. The face video was aligned by AAM and a combination of geometric features (similarity normalized shape points - S-PTS) and appearance features (canonical appearance - C-APP) was extracted. A linear SVM was used to perform binary classification of "pain" vs. "no-pain" , and prediction of pain score at frame level was defined as the distance of the test observation from the separating hyperplane of SVM. Pain scores for every frame are summed up and normalized for the duration of the sequence to produce a cumulative score for sequence level pain measure, and the decision threshold is determined by Equal Error Rate (\emph{EER}) derived from ROC. In the follow-up research Lucey \emph{et al}\cite{lucey2011automatically} reported on recovering 3D pose from 2D AAM \cite{xiao2004real} and performing statistical analysis based on ground truth to reveal that patients suffering from pain displayed larger variance in head positions. A set of binary linear SVM classifiers was trained for $10$ pain related AUs to replace the single binary pain SVM classifier in the previous work \cite{ashraf2009painful}. Frame-level pain measurement was then derived from the output scores of AU detectors that were fused using linear logistical regression (LLR). A combination of S-PTS, S-APP and C-APP yields the best average AU recognition performance of $0.78$ measured by \emph{AUC} \cite{ashraf2009painful}, and the performance was further improved to $0.81$ by compressing the spatial information with a discrete cosine transform (DCT) \cite{lucey2008improving}. Continuous research based on this framework was conducted in \cite{lucey2012painful}, where OPI label was employed for sequence level pain evaluation to mimic the decision making of human observers. The OPI labels are segmented into three classes based on intensity (0-1, 2-3, and 4-5), and a one-versus-all binary SVM classifier was trained for each OPI group to generate a confusion matrix. The frame-level pain label assigned by the classifier model produced the highest probability score and a sequence level pain label is generated from all its member frames via a majority vote scheme. Although reasonable classification rate was obtained for no pain class (OPI 0-1), the system did not report desired performance to discriminate low pain level (OPI 2-3) from high pain level (OPI 4-5). Further findings from the testing results suggests that rigid head motions do not necessarily contribute to prediction of OPI intensities. This series of studies is valuable as it paves the way for APD research to replicate the job of a care-giver when monitoring the pain of a patient.

\subsection{Pain Intensity Estimation}
Hammel \emph{et al} \cite{hammal2012automatic} built a pain intensity classification system for the UNBC-McMaster dataset using the AAM-CAPP-SVM framework. A Log-Normal filter bank tuned to $15$ orientations and $7$ central frequencies was applied to the C-APP features to enhance the modeling of deepening and orientation change of appearance features that characterize pain expressions. The pain metric is measured with the PSPI scale and the intensities are categorized into $4$ classes ($0,1,2,\geq3$) and four linear SVM are trained with the Log-Normal filtered C-APP feature map for the four levels of pain intensity. The pain intensity is estimated on a frame-by-frame basis and the performance is evaluated by \emph{F1} for 5-fold ($91\%-96\%$) and leave-one-subject-out ($45\%-67\%$) cross-validation. Moderate to high consistency between manual and automatic PSPI pain intensity was measured by \emph{ICC} for the two cross-validation schemes. Irani \emph{et al} \cite{irani2015pain} adopted spatiotemporal oriented energy features to model facial muscular motions in a pain intensity detection framework on the same pain dataset. The face sequence was segmented and aligned using AAM, and then a face patch was further divided into $3$ regions based on the prior knowledge of location of pain expression muscular activities. Facial muscular activities were extracted at the frame level by steerable and separable energy filters that were implemented by a second derivative Gaussian followed by a Hilbert transform tuned to $4$ directions. A histogram was generated from pixel-based energy features as per directions and regions to form region-based energy descriptors. Vertical motion and horizontal motion were computed from the histogram of spatial features in up-down and left-right directions respectively. Frame-level motion features were summed in time domain to generate a final spatiotemporal descriptor for each region. Pain intensity was estimated by a weighted linear combination of the vertical and horizontal motion scores. Experimental results reveal an improvement in no-pain and weak-pain recognition in the work of \cite{hammal2012automatic}. Lundtoft \emph{et al} \cite{lundtoft2016spatiotemporal} modified the pain intensity framework in \cite{irani2015pain} by introducing
a super pixel approach for face region segmentation. They argued a better recognition rate for no pain category can be achieved by only exploiting vertical motions of the spatiotemporal descriptor from the middle region of the face.

\subsection{Pain Detection in Clinical Settings}
An encouraging clinical application of automated pain detection was reported in \cite{sikka2015automated}\cite{sikka2014facial}. This study applied CVML models to assess pediatric postoperative pain and demonstrated good-to-excellent classification accuracy and strong correlation with pain ratings from human observers (parents and nurses). In the data collection stage, a single camera recorded video of 50 youth, 5 to 18 years old, for transient and ongoing pain conditions during 3 study visits after laparoscopic appendectomy. Facial activities were recorded for 5 minutes as a measure of ongoing pain. Then the surgical site was manually pressed for two 10-second periods to stimulate transient pain, and video recorded. Pain ratings from patients' self-report, parents, and nurses were recorded independently in a 11-point Numerical Rating Scale (NRS), where self-report pain rates were considered as the subjective ground truth and time since surgery provided the objective ground truth alternatively. Video segments with NRS rating $\geq4$ are defined as trials with pain, while those with NRS ratings of $0$ are defined as trials without pain. The computer expression recognition toolbox (CERT) was employed to independently score $10$ pain-related single AUs, one smile-related AU combination, and rigid head motion in 3D (yaw, pitch and roll) per frame. Three statistics (mean, 75th percentile, and 25th percentile) were computed for each of the $14$ frame-level raw features across the duration of a pain event (ongoing and transient), which were comprised of a 42-dimensional feature vector for the pain event. Two linear regression models with L1 regularization are trained on the same set of feature vectors for binary pain classification and pain intensity estimation tasks.

Performance of a binary pain model was evaluated via a $10$-fold cross validation with two metrics, where \emph{AUC} (in the range of 0.84-0.94) demonstrates good-to-excellent detection rate and Cohen's $\kappa$ provides phycological measurement on categorical agreement between raters with consideration of chance agreement. Categorical agreement \added{evaluated for} transient and ongoing pain was fair to substantial ($\kappa = 0.36-0.61$) for the model trained on self-report ratings and substantial ($\kappa = 0.61-0.72$) for the model trained on objective ground truth, where the prediction of automated system was more consistent than human observers on ongoing pain according to the $\kappa$ measurement. The pain intensity estimation model was evaluated by the Pearson correlations metric (both within and across subjects) using a leave-one-subject-out cross validation scheme. The best performance was achieved when the PIE system was trained with objective ground truth for both pain conditions, and the within-subject correlation ($r=0.80-0.86$) was consistently higher than the correlation for all subjects ($r=0.55-0.59$). The CVML model performance was at least equivalently to or exceeded that of a nurse for both BPC and PIE tasks, but was only equivalent to that of parents in PIE task. The analysis of results from PIE also supports the tendency for nurses to underestimate pain severity \cite{williams2002facial} when compared with children's self-report, especially under the ongoing pain condition.  Model performances was not affected by including demographic information in the training features.  This research outlines general procedures to for applying a CVML model in clinical settings, which involve experimental design and data collection, CVML tools, model evaluation from both engineering and phycology aspects, as well as performance comparison with human observers. Therefore the \added{methodology in this research} serves as an important modality for future APD research.

\subsection{Detecting Pain Event with 'Weakly Labelled' Data}
A patient's self-report is the golden rule for pain evaluation in patient care. A pain label is commonly available for a video sequence but not for every single frame. Such a situation is encountered frequently in computer vision since it is easier to obtain group labels for the data rather than individual labels, and is known as ’weakly supervised’ learning problem. Sikka \emph{et al} \cite{sikka2014classification} modeled a video sequence with a bag of words (BOW) structure that is comprised of multiple segments to handle the 'weakly labelled data', and solved the 'weakly supervised' pain detection problem with a multiple instance learning framework (MIL). Two partitioning schemes, normalized cuts (Ncuts) and scanning window (scan-wind), were employed to generate multiple segments (referred to as instances) from a video sequence (referred to as bag). According to BOW definition, a positive (pain) bag contains at least one positive instance and a negative (no pain) bag contains only negative instances. The frame-level feature extraction learns a mapping from the pixel space to a $d$-dimensional feature vector. Choice for frame-level feature descriptor is highly flexible, the multi-scale dense SIFT (MSDF) with a 4-level spatial pyramid was selected in this work \cite{sikka2014classification}. The segment-level feature representation was then generated from all frames in the segment via a max pooling strategy, such that an instance in a bag is also represented by a $d$-dimensional vector. A Milboost \cite{zhang2006multiple} framework based on gradient boosting was used to learn the posterior probabilities for both bags and instances, which were directly related to pain event prediction on segment level or sequence level. The frame-level pain score was also derived from posterior probability of segments by assigning the maximum score among all the segments it belongs to. The frame level pain score can be used for pain localization in time domain. The proposed APD framework was trained and validated on the UNBC-McMaster dataset, where sequences with OPI rating $\geq3$ are selected as positive samples and those with OPI rating $=0$ are selected as negative samples. The binary pain detection was evaluated with \emph{EER} ($83.7\%$) in the ROC. The pain localization task was evaluated as pain intensity estimation on frame level with ground truth of PSPI, and the performance was evaluated by \emph{F1} socre ($.471$) and Spearman's rank correlation \cite{kendall1938new} ($.432$).

The MIL framework detects pain (pain vs. no pain) in a Multi-Instance-Classification (MIC) setting, where the sequence-level intensity labels (VAS,OPI) of pain are binarized. It is, therefore, not suitable for a multi-level classification/regression problem like the pain intensity detection task. Ruiz \emph{et al} embedded an ordinal structure of labels to the bag representation, which naturally corresponds to various rating or intensity estimation tasks. In the proposed Multi-Instance Ordinal Regression (MIOR) weakly supervised settings, a video sequence was modeled as a bag where instances are consecutive video frames in the sequence. The bag label belongs to an ordinal set and instance labels were treated as latent ordinal states, which were inferred from a learned mapping from feature space to the structural output space. A multi-instance dynamic ordinal random fields (MI-DORF) framework built upon the hidden conditional ordinal random fields (HCORF) \cite{kim2010hidden} framework was employed to handle the multiple instance learning (MIL) problem incorporating both ordinal and temporal data structures, where the temporal dynamics within the instances was encoded by transitions between ordinal latent states. The proposed MI-DORF framework was then applied to pain intensity estimation task on UNBC-McMaster dataset, where a total of $157$ sequences from $25$ subjects were selected based on the low-discrepancy criteria between sequence- and frame-level pain label. The sequence-level pain label in an ordinal scale (OPI label) between $0$ and $5$ was used to train the system, and the frame-level label in PSPI was also normalized to the same scale and used for results validation. The frame-level (instance) features are represented by $49$ fiducial points tracked by AAM. For a fair comparison with MIL in \cite{sikka2014classification}, this study followed a similar experimental setup as in \cite{sikka2014classification}, while the output probability for binary pain classification in MIC methods was also normalized to $0-5$ as a representation of pain intensity. The performance on sequence-level prediction was measured by \emph{PCC} ($0.67$), \emph{MAE} ($0.80$), \emph{ICC} ($0.66$), \emph{Accuracy} ($0.52$), and \emph{F1} ($0.34$),  which outperformed other MIL methods \cite{sikka2014classification} \cite{hsu2014augmented} as well as the baseline model \cite{kim2010hidden}. Furthermore, the proposed approach achieved a comparable performance on frame-level pain intensity estimation ($ICC=0.40$) using only sequence-level ground truth towards the state-of-the-art fully supervised (using frame level ground truth) Context-sensitive Dynamic Ordinal Regression method \cite{rudovic2015context} ($ICC=0.59-0.67$). This result suggests a good trade-off between system performance and labor intensive frame-level annotation.

\subsection{Automated Detection of Cancer Pain}
Study of pain other than acute pain, such as chronic pain caused by cancer, is an important yet almost unexplored area in the current state of APD research. A unique dataset was created by D. Wilkie \cite{wilkie1995facial}, containing videos of $43$ patients suffering from chronic pain caused by lung cancer. The videos are captured in natural settings in the subjects' homes. The patients were required to repeat a standard set of randomly ordered actions as instructed such as sit, stand up, walk, and recline, in a $10$-minute video with a camera focused on the face area to record their facial expressions. Each video was partitioned into $30$ equal-duration, $20$-second subsequence. The subsequences were reviewed and scored for 9 AUs, possibly occurring in combination, by three trained human FACS coders independently, and the results were entered in a scoresheet that served as a basis for ground truth. Pain was declared to occur in a video subsequence if at least two coders agreed with each other on the occurrence of a set of specific AU combinations listed on the scoresheet. The intensity of pain for the entire video was measured on the total number of subsequences that were associated with pain labels. In view of the fact of highly imbalanced data samples captured in spontaneous conditions and lacking available ground truth from patients' self-report, Chen \emph{et al} \cite{chen2012automated} employed a rule-based method to analyze pain from the Wilkie's dataset. The rules for AU identification were developed upon geometric features that are extracted from trajectories of a subset of fiducial points in an AAM shape model. Decision trees were used to encode AU combinations and pain was identified from AU combination scores.

Most existing video datasets containing pain-related facial expressions are developed for targeted studies. These
datasets are typically small in size without public access, and lack sufficient diversity to train a robust
APD system in general. Chen \emph{et al} \cite{chen2017learning} decoupled an APD framework into a cascade of two machine learning modules: an AFER system that computes frame-level confidence scores for single AUs and a MIL system that performs sequence-level pain prediction based on contributions from a pain-relevant set of AU combinations. The AFER system was implemented with a commercial Emotient system (formerly the CERT) to handle general environmental challenges for spontaneous facial expression recognition. AU combinations are described by two distinct low-dimensional novel feature structures: (i) a compact structure encodes all AU combinations of interest into a single vector, which is then analyzed in the MIL framework \cite{zhang2006multiple}; (ii) a clustered structure uses a sparse representation to encode each AU combination separately followed by analysis in the multiple clustered instance learning (MCIL) framework \cite{xu2014weakly}, which is an extension of MIL. The 'weakly supervised' binary pain detection task was performed on UNBC-McMaster dataset and evaluated in terms of \emph{Accuracy} ($86.8\%$) and \emph{AUC} ($0.94$). A trans-dataset validation by applying the trained system (on UNBC-McMaster) to the Wilkie's dataset also demontrates good consistency between system prediction and human coders' decisions on both pain ($82.9\%$, $2+$ coders agreed) and no pain ($88.9\%$, three coders agreed) cases. This study suggest a potential way to solve the insufficiency of pain-labelled samples by fusing data from multiple small-scale pain datasets.

\subsection{Multi-Modality in Pain Detection}
Pain is a subjective experience, which is possibly caused by various stimuli (e.g., spinal stenosis, rotator cuff tear, cardiac conditions or sickle-cell disease) and manifested through signals from multiple modalities including voice, facial and body expressions \cite{hammal2014towards}. Present efforts in APD largely focused on acute pain that is elicited by a single stimulus in a controlled environment, which poses a gap between existing APD research and requirements of practical pain-related clinical applications. The UNBC-McMaster is the only publicly available dataset with well-annotated data for acute pain that has facilitated APD research since its first availability in 2011. Although it has been widely employed to benchmark the performance of both AFER and APD systems, its contribution is limited by its inability to lend itself to study other types of pain such as chronic pain.

A recent Emotion and Pain project extends the option for chronic pain analysis by introducing a fully labeled multi-modal dataset (named EmoPain) \cite{aung2016automatic} targeting at chronic lower back pain (CLBP). The data collected in EmoPain contains multiple signals with and across modalities, including high-resolution multi-view face videos, full body 3D motion capture, head-mounted and room audio signals, and electromyography signals from back muscles. Participants are requested to perform both instructed and non-instructed exercise to simulate therapist-directed and home-based-self-directed therapy scenarios. Level of pain from facial expressions was labelled by $8$ raters, while six pain-related body behaviors were annotated by $4$ experts. $22$ CLBP patients together with $28$ healthy control subjects were recruited as final participants for the CLBP study. The baseline experiment on facial expressions of pain used full length of $34$ unsegmented trials captured by frontal-view camera from $17$ patients, which are comprised of a total of $317,352$ frames ($33.3\%$ contained pain). The baseline system tracked $49$ inner facial landmarks that were aligned by a similarity transform. Two local appearance descriptors, the local binary pattern (LBP) and discrete cosine transform (DCT), are employed to extract features from local patches with a radius of 10 pixels centered on $30$ fiducial points. The two sets of appearance features together with geometric features from the landmark points are fed separately into a linear SVM for binary pain recognition task. The best performance was archived by the geometric feature, which is measured by \emph{AUC} ($.658$) and \emph{F1} ($.446$). Given facial expression as a communicative modality and body behavior as both communicative and protective \cite{sullivan2004communicative}, the author further performed a qualitative analysis of the relation between pain-related facial and body expressions. The results supported the finding that facial expressions of pain ($70\%$) occur in connection with a protective behavior, which may help the pain research community to better understand the relation among movement, exercise, and pain experience.

Wener \emph{et al} \cite{werner2013towards} addressed multi-modality pain detection by introducing the Bio Vid Heat Pain dataset, which include $90$ participants suffering from induced thermode pain in $4$ levels.  Multi-view face videos were recorded from three synchronized optical cameras in conjunction with depth map from a Kinect sensor. The physiological signals are captured from the skin conductance level (SCL), the electrocardiogram (ECG), the electromyogram (EMG), and the electroencephalogram (EEG). The pain detection experiment was based on only the video data in this study, where data analysis based on physiological data would be conducted in future research. A 6D (3D position and 3D orientation) head pose vector was computed from a 3D pointed cloud recovered from the Kinect depth map. The face expression was uncoupled from rigid motion by projecting the tracked fiducial points (from 2D video frames) via a pinhole camera model onto the surface of a generic face model in 3D according to the current head pose. Geometric features are defined by $8$ distance measure from the fiducial points. Some texture changes, including nasal wrinkles, nasolabial furrows, and eye closure are synthesized by mean gradient magnitudes extracted from local rectangular regions centered at $5$ selected fiducial points. The final feature descriptor per frame was in 13D, which is a combination of geometric and appearance features. The temporal features are extracted from the 6D head pose vector and 13D facial expression feature vector within a $6s$ temporal window. This resulted in a 21D descriptor per signal defined by 7 statistic measures (mean, median, range, standard and median absolute deviation, interquartile and interdecile range) for each of the smoothed signal and its first and second derivatives. A binary SVM classifier with radial basis function (RBF) kernel to discriminate pain from no pain was trained for each of the $4$ pain levels, and the performance was measured by the F1 metric. Further experiments on the highest pain level data suggested that include head pose features would improve system performance on both person-specific and general pain classifiers. However this finding may be dataset-dependent as a contradictory conclusion was reported by studies \cite{lucey2011automatically} using UNBC-McMaster dataset. A literature survey on machine-based pain assessment for infants \cite{zamzmi2016machine} reviewed progress on multimodal tools including behavior-based approaches (e.g. facial expression, crying sound, body movements) and physiological-based approaches (e.g. vital signs, and brain dynamics). However this is outside the scope of our review that is focused on pain experienced primarily by adult patients.

%
%case study
%Test subject
%Pain stimuli
%At which scale is pain measured( Frame Sequence Event)
%Feature collection
%Temporal Dynamic Modeling
%Direct way, indirect way
%Performance evaluation
%Face detection and alignment is less focused.

\section{Evolution with Deep Learning}

\subsection{AFER with Deep Learning }
Recent advances in deep learning techniques, especially the convolutional neural network (CNN), have provided an 'end-to-end' solution to learn facial action unit labels directly from raw input images, which significantly reduced the dependency on designing functional modules in a conventional CVML framework. The high-level deep features are learned from mass-scale simple features through a pipeline architecture which facilitates developing pose-invariant applications for AU detection. In fact, spontaneous facial expression datasets with large rigid motions that are challenging to conventional AFER studies, for example BP4D and DISFA, have been widely employed in recent deep learning based research. On the other hand, the structure of multi-class outputs in a deep neural network (DNN) can be conveniently modified at the output layer to handle a multi-label AU detection problem. In this case, AUs are jointly predicted with consideration on their co-occurrence and dependencies that is insufficiently exploited in previous one-versus-all binary classification settings. However, training a DNN from scratch relies on large labeled training datasets (100K+ samples in \cite{parkhi2015deep} ), which is generally not available in most facial expression datasets due to the labor-intensive FACS coding procedure. Therefore, deep learning based AFER with partially labelled data has also brought attention to recent studies. The literature review in this section will focus on the three problems presented above as these are important common problems to APD studies with deep learning.

\subsubsection {AU Recognition Model with a Deep Learning Architecture}
A deep learning based system generally follows a different paradigm than a conventional CVML system in performing the AU detection task, which focuses on three problems involving pre-processing, the DNN configuration and post-processing. The input images are cropped, normalized and aligned with image processing and computer vision tools so that they have a uniform format when applied to the input layer of a DNN. The DNN combines the feature extraction and classification module into a single network, where the architecture, training scheme, and optimization settings need to be specified. Finally the post-processing step determines how the raw data from the output layer are interpreted to the targeted AUs.

An example of the deep learning based framework \cite{gudi2015deep} for FACS AU occurrence and intensity estimation was proposed in FG 2015 Facial Expression Recognition and Analysis challenge (FERA 2015). The core architecture of the 7-layer CNN was composed of $3$ convolutional layers and a max-pooling layer as feature descriptor and a fully connected layer to provide classification output. In the pre-processing step, faces were detected, aligned and cropped into fixed size from input images followed by a global contrast normalization before they were fed to the CNN. The raw output values from the activation of the output layer were used as confidence scores to determine AU occurrence, where the optimal decision threshold was learned from a validation set based on highest \emph{F1} measure. The proposed deep learning AFER system was trained from scratch on both BP4D and SEMAINE datasets, where the performance for binary AU detection was measured by \emph{F1} ($0.522$ for BP4D, $0.341$ for SEMAINE) and the intensity estimation was evaluated on BP4D only with \emph{MSE} ($1.181$), \emph{PCC} ($0.621$) and \emph{ICC} ($0.613$).

A more extensive combination of features and learning framework was carried out \cite{tHoser2016deep} for frame-level AU detection under large head pose variations. Three type of features were extracted in the pre-processing step, including histogram of oriented gradients (HOG), similarity normalized facial images and a mosaic representation (cut and put together from patches around landmark positions).  The AU detection was formulated as a single- or multi-label classification problem respectively. Four types of network settings based on SVM and a customized small-scale CNN are investigated: 1) a single-label SVM with HOG features, 2) a single-label CNN with the normalized images, 3) a multi-label CNN with the normalized images, and 4) a multi-label CNN with the mosaic features. All the system are trained on the BP4D dataset from the FERA 2015 challenge and tested on an augmented dataset that is created using 3D information and renderings of faces from BP4D with large pose variations. The performance was evaluated for binary detection task of 11 AUs with \emph{F1} score, skew-normalized \emph{F1} score, and AUC. In general, the single-label CNN took similarity normalized features yielded the best performance over all four framework settings. Further testing with large head pose variation ($0^\circ$ to $-72^\circ$ for yaw and $-36^\circ$ to $36^\circ$ for pitch) indicated F1 score is a weak function of pose, which suggests that the deep-learning technique is capable of handling facial data under large pose variations.

Zhou \emph{et al} addressed the AU intensity estimation under large pose variations by developing a multi-task deep network \cite{zhou2017pose} for the sub-challenge of FERA 2017. They investigated a pose-invariant model and a pose-dependant model, which were both built upon the bottom five layers of a pre-trained VGG16 network \cite{simonyan2014very} via transfer learning as the feature descriptor. A two-layer fully connected regressor was trained for each AU independently in the pose-invariant model using data from one AU across all poses as the AU intensity estimator. In the pose-dependent model, the data of each AU were categorized into three groups based on the pitch pose. Three pose-dependent regressors were trained jointly with an auxiliary pose estimator under the assumption that the output of pose estimator is equal to the pose ground truth, and the final AU estimates were made from the dot product between the estimated pose vector and the three pose-dependent AU intensity estimates. Both models were trained on the BP4D datasets, and evaluated with the metrics of \emph{RMSE}, \emph{PCC} and \emph{ICC}.  The pose-invariant models using appearance features significantly outperform the baseline model in \cite{valstar2017fera} that relies on geometric features. Two likely reasons that the performance is affected are (i) the high failure rate on tracking geometric features under extreme pose change, and (ii) distortion to the facial structure introduced by the warping and alignment operation to the tracked landmarks. Comparison between the pose-dependent model and the pose-invariant model in terms of \emph{ICC} revealed improvement on AU 1, 4, 12, and 17, but no improvement on AU 6, 10, and 14. Hence a mixed model (pose-invariant or pose-dependent) can be selected for each specific AU to further improve the performance.

\subsubsection{Multi-label AU Detection}
In practice, AUs frequently occur in combinations so that a single image frame may carry positive labels from multiple AUs. The one-versus-all binary classification in conventional AFER studies either perform region-specific AU detection or ignored the dependencies among co-occurring AUs. As a result AUs are detected independently in most cases and certain AU combinations are also treated as a single variable for classification. On the other hand, the architecture of a DNN is naturally designed for multi-classification problem, which can be conveniently converted to handle multi-label learning with a proper loss function (e.g. multi-class sigmoid entropy), such that the sum of probabilities from all the output classes are not necessarily equal to one. The multi-label setting combined with deep learning provide improvement in handling imbalanced samples in many spontaneous facial expression datasets. Furthermore, the joint estimation of multiple AUs provide high flexibility in describing various AU combinations, which provide advantages in encoding high-level event like pain or emotions.

The deep region and multi-label (DRML) \cite{zhao2016deep} framework solved the problem of region-based feature extraction and joint estimation of multiple AUs in a unifying deep learning network. The core architecture of DRML was a CNN comprised of five convolutional layers and one max-pooling layer, and the multi-label option was enabled by configuring the output layer as a multi-label sigmoid cross-entropy loss. An additional region layer was attached after the first convolutional layer and partitioned the feature map from the convolutional layer with a fixed $8$ by $8$ grid. Each sub-region was processed by the same sub-network architecture formed by a batch normalization (BN) layer with ReLU activation and a convolutional layer. The coefficients of each sub-network in the region layer were updated independently with respect to local features and AU correlations, and the output of the region layer is a concatenation of all re-weighted patches. However, this settings requires good face alignment such that local features can be consistently captured in the same sub-regions.  The entire network was trained from scratch with data from BP4D (12 AUs) and DISFA (8 AUs) respectively and the performance was evaluated with \emph{F1} ($0.483$ on BP4D, $0.267$ on DISFA) and \emph{AUC} ($0.560$ on BP4D, $0.523$ on DISFA). The experimental results indicated that the detection of AUs with higher skewness benefited from multi-label learning, and the region layer also improved the detection rate of most AUs by better exploiting structural information from local facial regions.

Region segmentation paradigm based on geometric features in conventional AFER can also be extended to deep learning based approach. Li \emph{et al} \cite{li2017action} built a deep learning frame-level binary AU detection framework for region of interest (ROI) adaptation, multi-label, and LSTM based temporal fusing. Twenty regions of interest ($14\times14$) related to local muscular motions that activate corresponding AUs were generated from and centered around 20 fiducial points derived from the shape model. A VGG16 network was used as the base feature descriptor, where the output from $12$th convolutional layer ($224\times224$) was chosen as the feature map and cropped by the pre-defined sub-regions. A CNN-based ROI cropping net (ROI Net) was trained for each individual sub-region as an AU-specific feature descriptor. The AU detection was formulated as a multi-label problem at the output layer to jointly predict all AUs per frame. The static predictions were then fed into a LSTM network to refine the AU estimates using temporal information from the past $24$ frames. The proposed framework was trained for $12$ AUs in the BP4D dataset and and tested on both BP4D and DISFA datasets. Experimental results demonstrated contributions from all three aspects of region cropping, multi-label, as well as LSTM. The settings with multi-label ROI nets followed by a single layer LSTM yielded the best performance based on \emph{F1} score ($0.661$ on BP4D, $0.523$ on DISFA). The performance improvement over existing DRML approach was attributed to the adaptive way of sub-region selection and the adoption of a deeper pre-trained network (VGG) as base feature descriptor.

Walecki \emph{et al} \cite{walecki2017deep} integrated a conditional graph model to the output of a CNN to learn the the co-occurrence structure of AU intensity levels from deep features in a statistical manner. The CNN was comprised of 3 convolutional layers each with RELU activation followed by max-pooling, which was used as a feature descriptor to extract deep features from input image frames. Each image frame in the dataset carried a label vector of all AUs of interest and each entry in a label vector assumed a value from a six-point ordinal scale. The ordinal representation of  AU intensities and their pairwise co-occurrence relations were defined by unary (ordinal) and binary (copula) cliques in the output graph that was learned by a Conditional Random Fields (CRF) model. The CRF framework estimated the posterior probability of a label vector conditioned on the deep features in the form of a normalized energy function, which was defined by a set of unary and pairwise potential functions. The parameters of CNN, unary, and pairwise potential functions were then jointly optimized through a 3-step iterative balanced batch learning algorithm. A data augmentation learning approach was also applied to leverage information from multiple datasets, such that a single 'robust' model is trained across multiple datasets instead of training an individual 'weak' model on each dataset. Under the data-augmentation settings, the CNN was trained using data from multiple datasets simultaneously. However different CRF pairwise connections for AU dependencies were learned seperately for each dataset, as these datasets may contain non-overlapping AUs with considerable variation in dynamics. The training process of the frame-level based CRF-CNN AU intensity framework involves data from BP4D in FERA 2015, DISFA and UNBC-McMaster dataset, where the UNBC-McMaster dataset was only used for data augmentation purpose. The settings of CRF-CNN with iterative balanced batch approach and data-augmentation option achieved the best performance, which was measured by \emph{ICC} ($0.63$ on BP4D, $0.45$ on DISFA) and \emph{MAE} ($1.23$ on BP4D, $0.61$ on DISFA). The DISFA dataset benefited from the augmentation with $7\%$ improvement on ICC.

\subsubsection{AU Detection with Partially Labeled Data}
In general, FACS-based data annotation is far more labor intensive than facial expression data acquisition, such that a complete label assignment to all the training images in a dataset is not always available. By assuming intra-class similarity on feature extraction or a distribution depicting the dependencies among AUs, labels of unannotated images may be inferred using the set of labeled data. Learning with partially labeled data could enrich the feature space by improving utilization of all available data, which is especially beneficial to deep learning oriented approaches requiring training with extensive data.

In addition to the binary label set $\{-1,1\}$ that indicates negative samples and positive samples, a $0$ label is  assigned to an unlabeled sample in the partial label learning problem. Wu \emph{et al} proposed a multi-label learning with missing labels (MLML) framework \cite{wu2015multi} to solve this problem in a multi-label setting, which is applicable to AU recognition. The goal of this method is to learn a mapping that is specified by a set of coefficients between the extracted frame-level feature and the predicted label. The missing labels were handled based on two local label assumptions.  An instance-level smoothness assumes that two frames with similar features should have label vectors that are close, and a class-level smoothness assumed that two classes with close semantic meanings should have similar instantiations across the whole dataset. The consistency between predicted labels and provided labels was enforced the MLML objective functions, which is comprised of a Hinge loss and constraints from two smooth matrices generated upon the label smoothness assumptions. The AU classification task was tested on the CK+ dataset with the available portion of labels on four levels ranging from $20\%$ to $100\%$. The performance was evaluated by average precision (\emph{PR}) ($0.81 - 0.92$) and \emph{AUC} ($0.892-0.949$).

The label smoothness assumption in \cite{wu2015multi} could be invalid as the closeness of two samples in feature space may arise because the subject is same rather than indication of the same AU. Wu \emph{et al} argued that regular spatial and temporal patterns embedded in AU labels can be described probabilistically as a label distribution, and AU labels including the missing ones can be modeled as samples from such underlying AU distributions \cite{wu2017deep}. In their method, a deep neural network was used as a feature descriptor to extract high-level features from a raw input image, where only the top hidden layer and the output layer were involved in learning AU label distributions and AU classifiers. A Restricted Boltzmann Machine (RBM) was employed to learn AU distributions with respect to the partially annotated ground truth labels from the bias of AU labels in the output layer, bias of the hidden layer, and the weights between the output layer and the hidden layer. More specifically, the weights between the two layers were assumed to capture the mutually exclusive and co-occurrence dependencies among multiple AUs. Multiple SVMs were used as classifiers for multiple AU classification based on the high-level features. The object loss function was a Hinge loss constrained by the log-likelihood of the learnt AU label distributions. The loss function was modified such that all data are utilized in the training process but only the annotated data contributed to the loss optimization. The two-step procedure first trained the DNN with the partially annotated data in an unsupervised manner and then fine tuned the parameters of both DNN and SVM with respect to the object function. The system was first trained and tested with fully annotated data from BP4D and DISFA respectively and minor improvement was reported on the proposed system toward other state-of-the-art methods (e.g. DRML) in term of \emph{F1} measure. Further test with partially labeled data also demonstrated the proposed system outperformed MLML with missing data ratio ranging from $0.1$ to $0.5$.

%deep learning is usually used as base level feature descriptor.
%
%multi-labling
%partial label missing
%rigid motions
%feature representation
%
%preprocessing
%alignment
%feature selection
%classification
%Temporal Dynamic Modeling
%Multi-level Network
%Fusing hand-craft feature and deep features

\subsection{Applications of Deep Learning in Pain Detection}
Deep learning techniques have also been adopted to recent APD research, with focus on improving feature representation and exploiting temporal dynamics by utilizing corresponding deep neural network architectures. To our best knowledge, the handful of studies on APD involving deep learning are all based on UNBC-McMaster dataset.

An end-to-end deep learning pain detector presented in \cite{rudovic2015context} is an upgrade of the classic pain detection framework that is based on hand-crafted features. The proposed two-stage deep learning framework is comprised of a convolutional neural network (CNN) and long short-term memory network (LSTM). The CNN took raw images as input and predicted pain at frame level, which was built upon a pre-trained VGG-16 CNN and served as a baseline system. The feature frames ($4096$ dimensions) from the fully connected layer (fc6) before the output layer of the CNN were grouped into sequences of length $17$, and all the frames are assigned the label from the preceding frame in the sequence. The frame-level features were fed into the LSTM network to improve the pain prediction of a frame by taking into account the past $16$ frames. The input images were preprocessed in several settings including: 1) aligned with generalized Procrustes Analysis followed by cropping to a fixed size, 2) removing background with a binary mask, and 3) frontalization with down sampled shape model (C-APP). The proposed system was trained on UNBC-McMaster dataset with frame-level pain labels, and the best performance on binary pain detection task ($Accuracy=90.3\%$, $AUC = 0.913$ was reported by the Aligned Crop scheme combined with exploiting temporal information using LSTM. The author argued that removing background information caused worse performance as the zero background value would be converted to nonzero in training CNN. Further test with skew-normalization on the dataset showed that the Accuracy measurement varied significantly with skew in testing data while AUC was less affected, which is in accordance with the findings in \cite{jeni2013facing}.

One obstacle in applying deep learning to APD problem is that the size of a pain-oriented dataset is too limited to support training a deep neural network from scratch. Egede \emph{et al} \cite{egede2017fusing} addressed this issue by combining hand-crafted and deep-learned features to estimate pain intensity as a multi-modality fusing problem that is tailored to small sample settings on the UNBC-McMaster dataset. The deep learned features were obtained from two pre-trained CNN models for AU detection task on BP4D dataset \cite{jaiswal2016deep}, which are associated to the eye and mouth region of the face respectively. A difference descriptor was applied to the local patch sequences with a temporal window of $5$ frames to obtain frame-level features in the form of image intensity and binary mask, which were combined and fed as input of CNN. The output of the last fully connected convolution layer ($3072D$) of each CNN was retrieved and combined into a ($6144D$) deep-learned feature vector. In addition, a $218D$ geometric feature vector extracted from the $49$-points shape model and $2376D$ HOG feature vectors extracted from local patches around facial points are employed as hand-crafted features at frame-level. A relevance vector regressor (RVR) was trained for each of the geometric, HOG, and deep features, and a seconde-level RVR was trained to fuse the RVR predictions from each modality in the first level. All RVR learners are trained on the frame level ground truth annotated by PSPI. System performance was evaluated by \emph{RMSE} ($0.99$) and \emph{PCC} ($0.67$), with the most significant contributions from HOG, followed by geometric feature and least from deep learned feature. The author attribute these findings to the likelihood that the deep features were fine-tuned to the original AU detection problem in \cite{jaiswal2016deep}, whereas the hand-crafted features were more generic in nature. The author further concluded that MSE effectively measures the difference between the predictions and the ground truth, while Pearson's correlation measures how well the prediction follows the ground truth trend. Although a good system performance should possess low RMSE and high Pearson's correlation at the same time, a good rating on either metric is not necessarily correlated to the same good value on the other.

Liu \emph{et al} \cite{liu2017deepfacelift} proposed a direct estimation framework based on the UNBC-McMaster dataset to predict the subjective self-report pain intensity score represented by VAS via hand-crafted personal features and multi-task learning. The proposed DeepFaceLIFT (LIFT = Learning Important Features) framework is a two-stage personalized model, which is comprised of a 4-Layer neural network (NN) with ReLU activation, and Gaussian process (GP) regression models. A fully connected NN was trained on frame-level AAM facial landmark features to predict frame-level VAS score in a weakly supervised manner, where all frames within a sequence carried the sequence VAS label. Personal features (complexion, age, and gender) are also involved in the training stage via three separate settings (Not used, Appended to the 3rd layer of NN, Appended to Input Features), and the multi-task learning was referred to the training data labeled with VAS, and labeled with both VAS and OPI. Instead using a temporal model as in typical APD settings, the authors solved the sequence-level pain estimation as a static problem based on sequence-level statistics. A set of ten statistics (mean, median, minimum, maximum, variance, 3rd moment, 4th moment, 5th moment, sum of all values, interquartile range) was generated from frame-level predictions of pain score, as these \emph{sufficient statistics} \cite{hogg1995introduction} capture important information that help to infer unknown parameters in distribution models. The ten-dimensional statistical sequence-level feature vectors were fed into a GP regression model to obatian a personalized estimate of VAS on sequence-level. The system performance was evaluated by \emph{MAE} ($2.18$) and \emph{ICC} ($0.35$), when the first stage NN was trained with both VAS and OPI label in conjunction with personal features appended to the 3rd NN layer, and the second stage GP input used only the predicted VAS feature. This result suggests that using OPI scores from an external observer contribute to estimation of the subjective VAS score by exploiting the benefits of multi-task learning.

\section{Discussion and Future Work}
Advances in FACS-based automated pain detection can be realized to a large extent from three aspects. Phycological studies provide theoretical basis for a modality to link the subjective experience of pain to the observation of facial expressions. Progress in computer vision and machine learning research will lead to technical implementation of pain detection from facial expressions by mining relevant features from raw input image frames. The system architectures proposed to solve the automated pain detection problem have to be trained and validated with data collected from experiments in clinical settings to demonstrate its effectiveness in practice. The APD problem can be viewed an extension of the AFER problem, not only because pain is associated to a set of facial action units and their combinations (e.g. PSPI), but also for the reason that AFER is the core of an APD framework. In this review we began with a description of a general AFER framework prior to addressing the APD problem due to the similarity and inheritance on the methodologies between the two problems. It is worth noting that in recent AFER studies, the face detection and alignment modules were emphasized less than the feature extraction and classification modules. This is likely due to the fact that the fiducial points have been well-annotated in most publicly available datasets, and state-of-the-art face detector and alignment algorithms are also available as convenient off-the-self toolboxes. Although choices of customized combination of geometric, appearance, and motion-based features are highly flexible, the SVM or boosting-based classifiers are consistently selected in conventional approaches, where single AUs are detected independently using a one-versus-all paradigm in general.  Dynamic approaches are often employed as additional steps to smooth the noisy static AU predictions at frame level by exploiting temporal information. Dynamic methods are also required in segment-level AU event encoding and localizing AU events by modeling the temporal phase transition.

Deep neural network has received increasing attention in AFER research since $2015$, which is able to provide an end-to-end solution for AU detection from raw input images. A DNN network possess an advantage over conventional machine learning classifiers in handling multi-class classification naturally, where the DNN architecture can be further modified to jointly detect action units as a multi-label problem. A deep neural network is able to synthesize high-level features from large-scale simple low-level features, and hence it is capable of substituting the conventional feature extraction module in an AFER framework with an improved modeling of the non-structural facial muscular movement. On the other hand, deep feature descriptors are often used in conjunction with other hand-crafted feature descriptors to enrich facial feature extractions in the current state of AFER research. Although training a DNN from scratch requires a large amount of data, which is not always available in a facial expression dataset, such a constraint may be alleviated via transfer learning from pre-trained state-of-the-art DNN models. However, a DNN framework behaves like a black box in feature selection in contrast to hand-crafted feature descriptors due to the complexity of its architecture eluding insightful analysis so far. To address this issue, a saliency map \cite{simonyan2013deep} that reflects gradient change at the input layer by disturbance at the output layer via back-propagation is used to visualize the active regions for feature extraction related to a particular AU.

The primary goal of automated pain detection research in clinical settings is to design an automated system that outperforms human observers in terms of both accuracy and efficiency on recognizing pain from facial expressions. The objective of pain detection is to encode segment-level events in the form of occurrence or intensity of pain. The ground truth of pain is available from patients' self-report at segment level in practice and identification of pain usually requires observation of co-occurring pain-related AUs. Therefore, clinical-oriented APD may be best described as a weakly supervised problem based on multi-label AU detection. Although development of an APD system fully benefits from progress in AFER performance as virtually any AFER techniques can be extended to APD applications, the reported studies on APD is are far fewer than those on AFER. A few studies using automated analysis of facial expressions in clinical settings \cite{hamm2011automated}\cite{sikka2015automated}\cite{liu2017deepfacelift} were more focused on psychological studies on interaction with human subjects rather than \added{elaborated} CVML implementations.  The establishment of the UNBC-McMaster Shoulder Pain Expression Archive Dataset has indeed motivated APD research since $2009$. However, the UNBC-McMaster dataset is still the only publicly available facial expression dataset about pain so far, which constrains most APD studies to the acute shoulder pain. It is worth noting that most present APD models are still in their infancy with limited consideration on psychological experiments. There is still a big gap between available APD models and the critical demand of pain monitoring applications for non-communicative patients. The major challenge impeding the progress on current APD research is primarily due to insufficient pain-annotated data for training and validating advanced CVML architectures. Aside from continuing evolution of CVML technology, the high priority goal in future work of APD will be collecting copious pain-annotated facial expression data by following well-designed protocols in clinical settings and making the new facial expression datasets publicly available. Mining AU relations from big data analysis will in turn help to optimize theoretical basis by providing more details on developing APD systems tailored to specific pain types.

\bibliographystyle{plain}
\bibliography{mybibfile}

% biographies
\newpage

\begin{IEEEbiography}[{\includegraphics[width=1in,height=1.25in,clip,keepaspectratio]{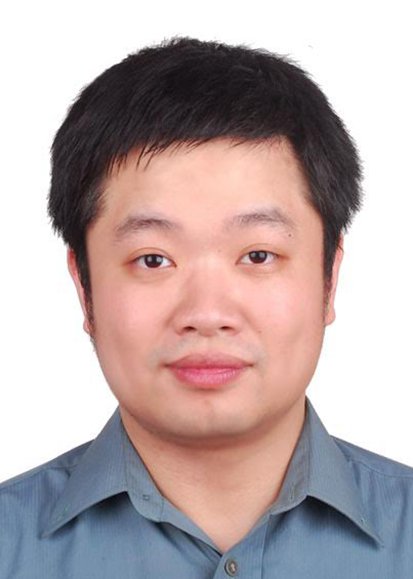}}]{Zhanli Chen}
is a Ph.D student in the Department of Electrical and Computer Engineering at the University of Illinois at Chicago (UIC). He received the Msc. degree in electrical engineering at Hong Kong University of Science and Technology (HKUST) and the B.Eng in electrical engineering at Nanjing University of Posts and Telecommunications (NUPT). He is a student member of IEEE and his research interest involves automated pain detection from facial expressions in clinical conditions.
\end{IEEEbiography}

\begin{IEEEbiography}[{\includegraphics[width=1in,height=1.25in,clip,keepaspectratio]{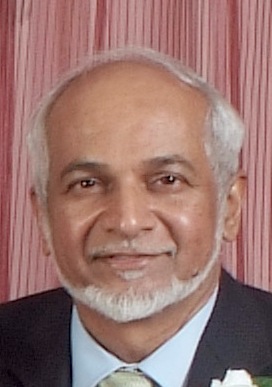}}]{Rashid Ansari}
received the Ph.D. degree in electrical engineering and computer science in 1981 from Princeton University, Princeton, NJ, and the B.Tech. and M.Tech. degrees in electrical engineering from the Indian Institute of Technology, Kanpur, India. He is currently Professor and Head in the Department of Electrical and Computer Engineering at the University of Illinois at Chicago (UIC). Before joining UIC he served as a Research Scientist at Bell Communications Research and on the faculty of Electrical Engineering at University of Pennsylvania. His research interests are in the general areas of signal processing and communications, with recent focus on image and video analysis, multimedia communication, and medical imaging. In the past he served as Associate Editor of the IEEE Transactions on Image Processing, IEEE Signal Processing Letters, and IEEE Transactions on Circuits and Systems.  He has served as a member of the Digital Signal Processing Technical Committee of the IEEE Circuits and Systems Society and a member of the Image, Video, and Multidimensional Signal Processing Technical Committee. He was a member of the organizing and program committees of several past IEEE conferences. He served on the organizing and executive committees of the Visual Communication; Image Processing (VCIP) conferences and was General Chair of the 1996 VCIP Conference. He was elected a Fellow of the IEEE in 1999.
\end{IEEEbiography}

% if you will not have a photo at all:
\begin{IEEEbiography}[{\includegraphics[width=1in,height=1.25in,clip,keepaspectratio]{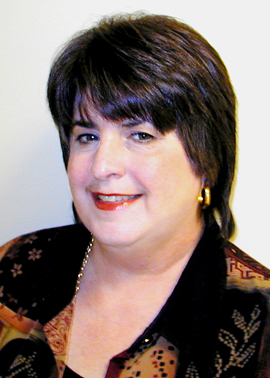}}]{Diana Wilkie} Ph.D., R.N., FAAN, comes to University of Florida from the University of Illinois at Chicago, where she served as professor and Harriet J. Werley Endowed Chair for Nursing Research. She joined UF in 2015. She is a fellow of the American Academy of Nursing and served for five years as an American Cancer Society professor of oncology nursing. She has authored or co-authored more than 210 scholarly publications and serves as a reviewer on a number of journals, including the Journal of Palliative Medicine, Cancer Nursing and the Journal of Pain. As a member of the National Academy of Medicine (formerly the Institute of Medicine), she has devoted her research program to management of cancer pain and to end-of-life issues. She has been continuously funded since 1986 from numerous organizations such as the National Institutes of Health, the National Cancer Institute and the Robert Wood Johnson Foundation — totaling more than \$48 million. She now directs the UF Center for Palliative Care Research and Education.

\end{IEEEbiography}

\end{document}